\documentclass{article}
\usepackage{spconf,amsmath,graphicx}

\usepackage{multirow} 

\usepackage{enumitem}
\setlist{nosep, leftmargin=14pt}

\usepackage{mwe} 

\usepackage{hyperref}
\usepackage{caption}
\title{A Comparison Of Object Detection And Phrase Grounding Models In Chest
X-Ray Abnormality Localization Using Eye-Tracking Data}
\name{Elham Ghelichkhan$^{1, 2}$, Tolga Tasdizen$^{1, 3}$ }
\address{$^1$Scientific Computing and Imaging Institute, University of Utah, Utah, USA\\ $^2$Kahlert School of Computing, University of Utah, Utah, USA\\ $^3$Department of Electrical and Computer Engineering, University of Utah, Utah, USA}

\begin{document}
%
\maketitle
\begin{abstract}
Chest diseases rank among the most prevalent and dangerous global health issues. Object detection and phrase grounding deep learning models interpret complex radiology data to assist healthcare professionals in diagnosis. Object detection locates abnormalities for classes, while phrase grounding locates abnormalities for textual descriptions. This paper investigates how text enhances abnormality localization in chest X-rays by comparing the performance and explainability of these two tasks. To establish an explainability benchmark, we proposed an automatic pipeline to generate image regions for report sentences using radiologists' eye-tracking data\footnote{The dataset is available at https://physionet.org/content/latte-cxr/}. The better performance - $mIoU=36\%\textit{ vs. }20\%$ - and explainability - Containment ratio $48\%\textit{ vs. }26\%$ - of the phrase grounding model infers the effectiveness of text in enhancing chest X-ray abnormality localization. 
\end{abstract}

\section{Introduction}\label{sec:intro}
Since the emergence of deep neural networks (DNN), they have been applied to various medical domains and applications. Chest diseases are the most common and dangerous health issues worldwide and chest X-ray (CXR) radiography is among the most frequently used and low-cost medical imaging modalities \cite{ait2023review}. Therefore, many works have applied DNNs to CXRs from the early stages of DNNs \cite{bar2015deep}, continuing to ongoing research today \cite{ait2023review, sharma2024systematic}. Despite their remarkable capabilities, there remains much room for improvement in performance and explainability. 

To enhance performance, researchers have integrated multiple modalities with radiographs, including eye-tracking (ET) data \cite{wang2024gazegnn, bigolin2023localization} and radiology report \cite{zhang2023knowledge}. Although the vast amount of clinical image-text pairs has facilitated the application of global multi-modal tasks, e.g., classification, image-to-text, and text-to-image retrieval \cite{zhang2022contrastive, huang2021gloria}, local visual-language models (VLMs) lack sufficient data in this domain.

VLMs, such as phrase grounding (PG), facilitate the interpretation of medical images and disease diagnosis by precisely aligning specific image regions with a disease description. However, aligning image regions and text is costly and time-intensive due to the need for expert annotators, precise annotation, and the complexities of medical diagnosis, leading to the scarcity of local VLM datasets. We proposed an automatic pipeline to generate a local image region, so-called bounding box (BB), aligned with a sentence of an image report, utilizing ET data from the REFLACX dataset \cite{reflacx, physioreflacx}. REFLACX radiographs are sampled from MIMIC-CXR \cite{johnson2019mimic1, johnson2019mimic2}, with only one frontal X-ray, present in the labels table of MIMIC-CXR-JPG \cite{johnson2019mimicjpg, DBLP:journals/corr/abs-1901-07042}.

\begin{figure}[t!]
\begin{minipage}[b]{1\linewidth}
  \centering
  \centerline{\includegraphics[trim={0 10cm 1cm 0},clip, width=8.5cm]{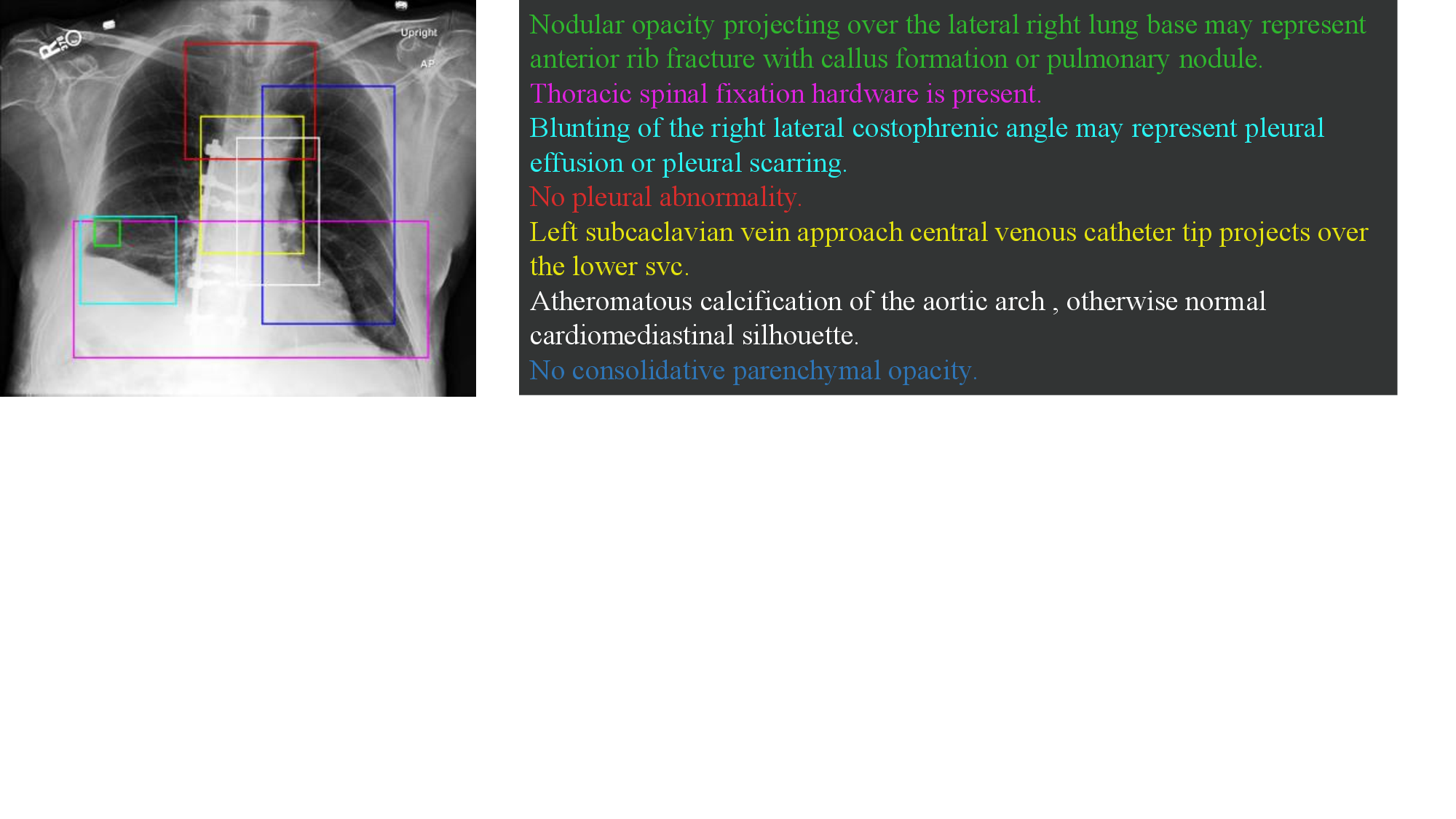}}
  \centerline{}\medskip
\end{minipage}
\caption{Automatically generated BBs for report sentences.} \label{met_dataset_example}
\end{figure} 

Previous works have integrated ET with vision-based and VLMs to improve classification \cite{wang2024gazegnn, zhu2022gaze, kholiavchenko2022gaze}, abnormality localization \cite{bigolin2023localization}, object detection \cite{hsieh2024eyexnet}, visual question answering, and report automation \cite{kim2024enhancing} on CXRs. We leveraged ET to automatically generate BBs for each sentence of a radiology report, introducing an evaluation metric and dataset for benchmarking explainability in chest X-ray abnormality localization models. To our knowledge, it is the first study to automatically generate bounding boxes corresponding to text using ET data in the medical domain. Applying this pipeline to the REFLACX dataset and validating by a phrase grounding  model  \cite{medRPG}, we demonstrated that the generated ET bounding boxes are learnable by a DNN ($mIoU \approx 0.36$) and informative, they contain $75\%$ of radiologists' annotations.  

While this pipeline efficiently provides data for VLMs, the advantage of VLMs over vision-based models still needs to be investigated. For instance, object detection (OD) and PG tasks aim to predict bounding boxes containing abnormality; one given an image, the other given an image and text. We repurposed REFLACX \cite{reflacx, physioreflacx} for PG and proved the effectiveness of a PG model \cite{medRPG} compared to YOLO-v5 \cite{yolov5} object detector, in nearly all classes. The PG model outperformed the OD model model on rare classes, when the OD model completely failed ($IoU\approx 0$). Further, by comparing the eye-tracking BBs with the models predictions, we observed that the PG outputs fall within ET bounding boxes more than OD, indicating that the PG model is more explainable. 

In the next section (\ref{methods}), we outline our approaches to repurpose an OD dataset (which also includes reports) for PG (\ref{met_repurpose}) and to automatically generate a PG dataset from image, time-stamped text, and ET data (\ref{met_dataset}). We also detail the localization methods we used in this study and compare their outputs with ET bounding boxes (\ref{met_pg_vs_od_vs_et}) with the appropriate evaluation metrics \ref{met_metrics}. Experiments (\ref{experiments}) presents and analyzes the results, and Discussion and Conclusion (\ref{discussion}) interprets the results, discusses the limitations, and concludes some future directions for this study.

\section{Methodology} \label{methods}
\subsection{Dataset Repurposing}\label{met_repurpose}
Each PG sample consists of an image, a BB, and a statement - textual description referring to the BB-, while each OD sample consists of an image, a BB, and a label. These can be written as triplets (image, BB, statement) and (image, BB, label) for PG and OD, respectively. Having both label and statement aligned with each BB, we compare PG and OD models. 

To gather a statement for an annotated BB from REFLACX \cite{reflacx, physioreflacx}, firstly, we removed sentences from reports which imply the absence of pathologies by filtering out sentences including term \textit{normal} and excluding term \textit{abnormal}, and sentences including \textit{'no '} (like \textit{No acute fracture.}). Then, concatenated the sentences that imply at least one of the BB labels; thus, a statement may contain more than one sentence. 

Some statements in report describe multiple BBs. The REFLACX BBs are annotated with certainty level of radiologists about the abnormalities (1 to 5). For a statement that corresponds to multiple BBs, we randomly chose one of the BBs with the highest certainty to pair with the statement.

\subsection{Automated Dataset Generation}\label{met_dataset}
\begin{figure}[t!]
\begin{minipage}[b]{1.0\linewidth}
  \centering
  \centerline{\includegraphics[trim={0 4.3mm 0 0},clip, width=7cm]{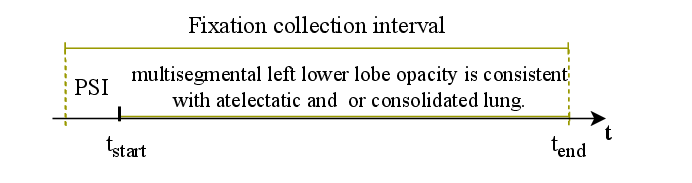}}
  \centerline{(a)}\medskip
\end{minipage}
\begin{minipage}[b]{.3\linewidth}
  \centering
  \centerline{\includegraphics[trim={3mm 8mm 3mm 7mm},clip, width=3.5cm]{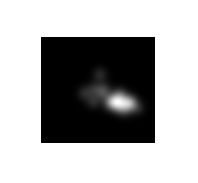}}
  \centerline{(b)}\medskip
\end{minipage}
\hfill
\begin{minipage}[b]{.3\linewidth}
  \centering
  \centerline{\includegraphics[trim={3mm 8mm 3mm 7mm},clip, width=3.5cm]{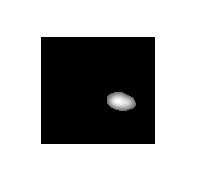}}
  \centerline{(c)}\medskip
\end{minipage}
\hfill
\begin{minipage}[b]{.3\linewidth}
  \centering
  \centerline{\includegraphics[trim={3mm 8mm 3mm 7mm},clip, width=3.5cm]{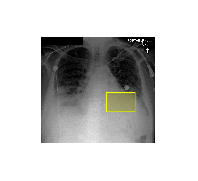}}
  \centerline{(d)}\medskip
\end{minipage}
\hfill
\caption{Automatic pipeline of ET bounding box generation. (a) Starting from the first sentence, we collected fixations from PSI seconds before the sentence starts until the sentence ends. (b) We summed all fixation heatmaps of a sentence to one heatmap; (c) normalized and thresholded them and removed small regions. (d) The BB enclosing the filtered heatmap is the extracted ET bounding box for the sentence.}  \label{met_AutomaticAnnotation}
\setlength{\belowcaptionskip}{-10pt} 
\end{figure}    
   
This section presents a pipeline to automatically generate data (Figure \ref{met_dataset_example}) for local VLMs using images, timestamped ET fixations, and timestamped medical reports. ET fixations are ET data gathered in the moments when a radiologist's gaze is stable over a specific region \cite{reflacx}. The timestamped ET and report were collected while a radiologist was interpreting a radiology image \cite{reflacx}. The following steps generate ET phrase grounding triplets by locating a BB for each report sentence.

\begin{enumerate}
    \item Starting from the first sentence, for every sentence, we collected the fixations from Pre-Sentence Interval (PSI) seconds before when the sentence starts ($t_{start}$) until when the sentence ends ($t_{end}$). Based on analysis performed in \cite{reflacx}, we set $PSI=1.5$. Then, we assigned each fixation to the first sentence that covers its duration (Figure \ref{met_AutomaticAnnotation}a).
    \item For each fixation, we generated a Gaussian heatmap proportional to the fixation duration, as in \cite{bigolin2023localization}. Then, we summed up heatmaps of all fixations corresponding to a sentence and normalized it to [0, 255] (Figure \ref{met_AutomaticAnnotation}b).
    \item Next, we thresholded the heatmap intensities by 40\% and removed objects smaller than $\frac{1}{400}\text{ image area}$ (Figure \ref{met_AutomaticAnnotation}c). 
    \item The rectangle enclosing the heatmap from the previous step is the BB corresponding to the sentence (Figure \ref{met_AutomaticAnnotation}d). 
\end{enumerate}
We performed experiments with different values to set the thresholds and visually optimized them to enclose the ET fixations. We will call these BBs as \textbf{ET bounding boxes}, and the ones enclosing abnormalities as \textbf{abnormality BBs}.

\subsection{Localization Methods}\label{met_pg_vs_od_vs_et}
Both OD and PG tasks predict image regions; one given a label and the other given a statement. A statement may contain information about location (\textit{right, left, central, lower, upper}), size (\textit{small, medium, large}), severity (\textit{mild, moderate, severe}), cause and effect relationships (\textit{ faint bibasilar parenchymal opacities may represent atelectasis or aspiration.}), confidence (\textit{probable, possible, likely, unlikely}), and evidence (\textit{multisegmental left lower lobe opacity is consistent with atelectatic and/or consolidated long}). We investigated the role of text in locating abnormalities by comparing MedRPG \cite{medRPG} and YOLO-v5 \cite{yolov5} performance and explainability. 

The data generated by our automatic pipeline differs from manually annotated data in some ways. Firstly, ET data is naturally noisy; different radiologists pay attention to different regions of the image, state different reports, and diagnose differently; applying the same thresholding setup to enclose their ET further deteriorates the issue. By comparing MedRPG \cite{medRPG} performance on ET and abnormality BBs, we will show that the ET bounding boxes are learnable by DNNs.

Secondly, since a larger image region is scanned to accurately locate an object, we expect the ET region to exceed the object's size. These larger scanned regions -recorded as ET fixations, enclosed in ET bounding boxes- contain the area the radiologists focused on to diagnose the abnormalities. Therefore, if a model prediction for a label or statement falls whitin the ET bounding boxes, it is explainable. Containment Ratio (CR) scores the model explainability by measuring the ratio of a prediction fallen in the corresponding ET bounding box.

\subsection{Evaluation Metrics}\label{met_metrics}
\begin{figure}[t!]
\begin{minipage}[b]{1\linewidth} 
  \centering
  \centerline{\includegraphics[trim={0 10cm 10cm 0},clip,width=8.5cm]{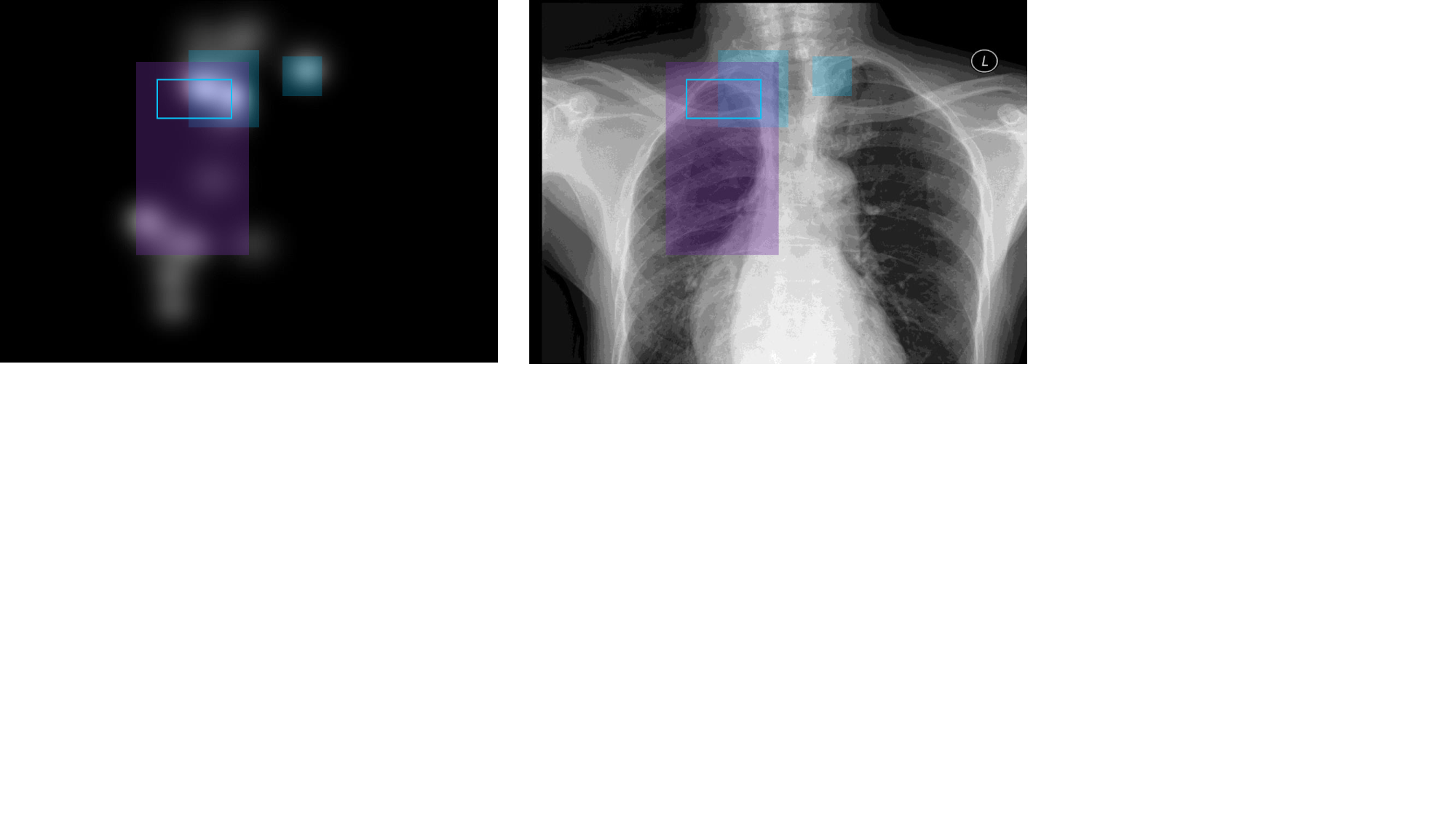}}
  \centerline{}
\end{minipage}
\caption{Explainability assessment by containment ratio. The blue-filled, purple-filled, and blue-outlined BBs show the abnormality annotations labeled as \textit{Pleural abnormality}, ET, and PG predicted BBs for statement \textit{biapical pleural thickening versus pleural fluid}, respectively. BBs are overlaid on the ET fixation heatmap corresponding to the statement (left).}\label{cr_for_explainability}
\end{figure}

We applied different evaluation metrics across different methods as follows: class-wise Intersection over Union (IoU), mean Intersection over Union (mIoU), and accuracy (Acc.) with IoU thresholds of 0.3 and 0.5 to assess performance; and containment ratio ($CR=\frac{\textbf{ET}\cap\textbf{abnormality BB}}{\textbf{abnormality BB}}$) to assess explainability. For instance, the explainability of the prediction in Figure \ref{cr_for_explainability} is $CR=1$ since it is completely covered by the ET BB (the informative region for radiologists); while for performance we calculate its IoU with the annotated abnormality BB, $IoU\approx 0.27$. It is worth mentioning that mIoU calculated in PG averages the score of the BBs, while mIoU calculated in OD averages over classes. To be comparable, the mIoU scores in Table \ref{tab_pg_od} are calculated in OD fashion.

\section{Experiments and Results}\label{experiments}
\subsection{Dataset}
We performed OD and PG on REFLACX \cite{physioreflacx} phase 3 data, where one of five radiologists annotated each image, including chest X-rays, reports, annotated ellipses, certainty levels, and ET fixations. We converted the ellipses to BBs, randomly assigned some training samples to validation to gain 80/20 for train/val, and created ET bounding boxes for the reports' sentences. For explainability, models were evaluated on REFLACX phase 2 data, annotated by all five radiologists. 

Both PG and OD models are trained and evaluated on phase 3 with 915/221/385 images for train/val/test. Each BB labeled by multiple classes forms multiple triplets for OD and one for PG (one statement in Figure \ref{cr_for_explainability} corresponds to two annotated BBs, we pair one of them with the statement); resulting in 1409/1998 training triplets for PG/OD models.

\subsection{Performance Comparison}\label{ex_pg_vs_od}
We trained YOLO-v5 \cite{yolov5} with manually annotated REFLACX \cite{physioreflacx} triplets and MedRPG \cite{medRPG} with PG repurposed triplets to predict abnormality BBs. Table \ref{'tab_pg_performance'} illustrates MedRPG performance on ET and abnormality BBs, with and without pretraining text encoder on Bio-clinical BERT \cite{biocli}. For the rest of our PG experiments, we utilized MedRPG with the pretrained text encoder. We trained and evaluated MedRPG using ET bounding boxes (test $mIoU=36.07$), demonstrating that despite their inherent noise (section \ref{met_pg_vs_od_vs_et}), these annotations remain learnable, thus validating their potential use for training DNNs.

\begin{table}
\caption{MedRPG performance on REFLACX abnormality (Abn.) and ET bounding boxes, Text encoder pretrained on BioClinical BERT \cite{biocli} or no pretraining}\label{'tab_pg_performance'} \centering
\begin{tabular}{|c|c|c|c|c|}
\hline
\multirow{2}{*}{BB} & \multirow{2}{*}{Pretrain} & \multirow{2}{*}{mIoU} & Accu. & Accu. \\
 & &  & \scriptsize \textit{$IoU \geq 50\%$} & \scriptsize \textit{$IoU \geq 30\%$}  \\
\hline
Abn. & Text Enc. & 50.52 & 52.00 & 76.89 \\
Abn.  & - & 45.4 & 49.86 & 66.67 \\
\hline
ET  & - & 36.07 &30.09 & 55.85 \\
\hline
\end{tabular}
\end{table}

\begin{table}[t!]
\caption{Comparison of PG and OD on REFLACX phase 3.}\label{tab_pg_od}\centering 
    \renewcommand{\arraystretch}{1} 
    \begin{tabular}{|l|c|c|c|}
        \hline
       Label & \#Labels  & OD (IoU) & PG (IoU)
        \\
        \hline
        \scriptsize Abnormal mediastinal contour & 11 & 9.19 & 3.94\\
        \scriptsize Acute fracture & 9 & 0 & 9.49\\
        \scriptsize Atelectasis &  189 & 11.26 & 32.1\\
        \scriptsize Consolidation & 171 & 14.12 & 31.44\\
        \scriptsize Enlarged cardiac silhouette & 165 & 73.16 & 77.49\\
        \scriptsize Enlarged hilum & 19 & 3.41 & 32.66\\
        \scriptsize Groundglass opacity & 99 & 0.22 & 29.34\\
        \scriptsize Hiatal hernia & 5 & 0 & 33.16\\
        \scriptsize High lung volume/emphysema & 28 & 29.48 & 30.81\\
        \scriptsize Interstitial lung disease & 7 & 0 & 22.17\\
        \scriptsize Lung nodule or mass & 34 & 5.71 & 18.92\\
        \scriptsize Pleural abnormality & 253 & 17.56 & 29.35\\
        \scriptsize Pneumothorax & 15 & 0 & 18.59\\
        \scriptsize Pulmonary edema & 179 & 5.35 & 28.24 \\
        \hline
        \multicolumn{2}{|c|}{\makebox[0pt][r]{\textbf{mIoU}}} & 19.61 & 36.20\\
        \hline
    \end{tabular}
\end{table}

\begin{figure}[t!]
\begin{minipage}[b]{1\linewidth} 
  \centering
  \centerline{\includegraphics[trim={0 0cm 10cm 0},clip,width=8.5cm]{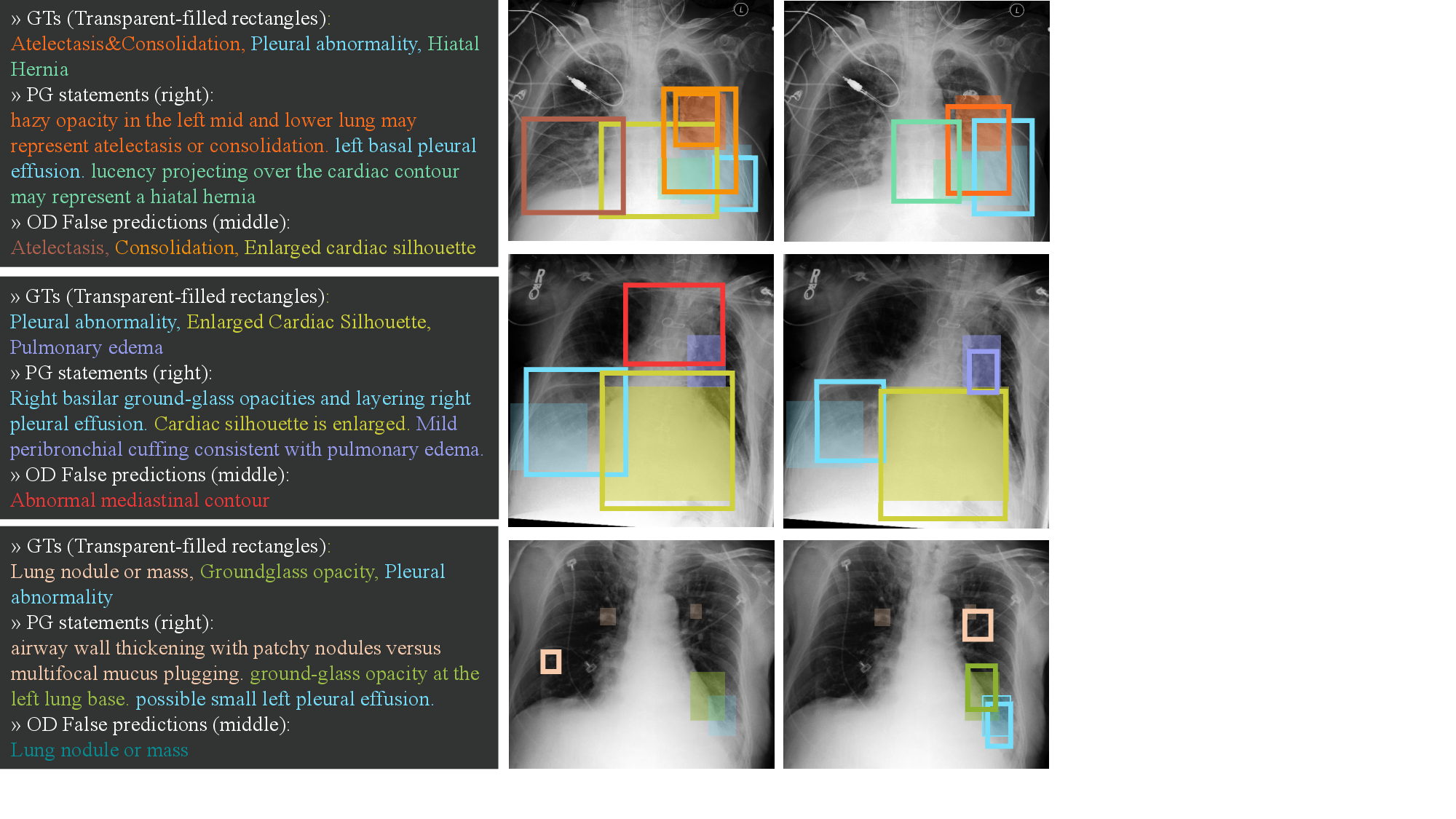}}
  \centerline{}
\end{minipage}
\caption{PG (right column) outperforms OD (middle column). The text boxes show color-coded GT, PG statements, and OD false predictions. Transparent-filled and outlined BBs show the color-coded GT and predicted BBs, respectively.}\label{pg_outperforms_od}
\end{figure}

Although our repurposed data can confuse training of the PG model (a statement may describe different BBs), the PG model outperforms the OD model, on average ($mIoU=36.2\% vs. 19.61\%$) and class-wise (Table \ref{tab_pg_od}). Both methods perform well ($IoU > 73\%$) on an easy class like \textit{Enlarged Cardiac Silhouette} with large BBs placed in roughly the same location throughout all images (Figure \ref{pg_outperforms_od} middle row) and perform weak ($IoU=0 vs. 9.49\%$ in OD vs. PG) on a challenging class like \textit{Acute Fracture} in ribs, with small and subtle pathologies. 

In contrast to the PG model, the OD model does not detect abnormalities ($mIoU\approx 0$) on classes with a low number of samples (\textit{Hiatal Hernia}, \textit{Interstitial Lung Disease}, \textit{Enlarged Hilum}, and \textit{Pneumothorax}). Two reasons can explain the better performance of PG: (1) a PG model deterministically generates a BB for each ground-truth BB and (2) a statement provides more information rather than a label (section \ref{met_pg_vs_od_vs_et}). 

In conclusion, YOLO-v5 \cite{yolov5} trained on a relatively small training set with 14 classes, performs weakly in most classes, while MedRPG \cite{medRPG} locates abnormalities more accurately by integrating textual and visual information. Figure \ref{pg_outperforms_od} displays annotated and predicted abnormality BBs. 

\subsection{Explainability Comparison}
This section uses REFLACX \cite{physioreflacx} phase2 data to analyze the explainability of YOLO-v5 \cite{yolov5} and MedRPG \cite{medRPG}, OD and PG models; by comparing their predictions with automatically generated (ground-truth) and predicted ET bounding boxes inferred from the MedRPG model, trained on phase3.

In order to assess the validity of ET bounding boxes, we compared automatically generated ET bounding boxes with annotated abnormality BBs. ET bounding boxes contain 74.84\% of abnormality BBs (Table \ref{'tab_CR'}). The rest of the abnormality BBs, not contained in ET bounding boxes, have been filtered out in the thresholding step of BB generation (section \ref{met_dataset}). Although ET data is naturally noisy, a high containment ratio shows ET bounding boxes contain salient image regions, aligned with the corresponding text and abnormality. This table also demonstrates that the containment ratio of the PG predictions is roughly twice of the OD predictions, indicating the MedRPG is more explainable than YOLO-v5. 

In addition, the high CR between predicted ET and predicted abnormality BBs, compared to ground-truth ET bounding boxes and predicted abnormality BBs, reveals two facts about bias in annotations and predictions. Firstly, the predictions are inferred by models trained on all radiologists' annotations; they are less biased than one radiologist's annotation. Secondly, both predicted ET and predicted abnormality BBs are affected by the same algorithmic bias in MedRPG.

\begin{table}[t!]
\caption{Containment Ratio of abnormality (Abn.) BBs in ET bounding boxes. GT (ground-truth) eye-tracking BBs are the automatically generated ones. Pred: prediction.}\centering\label{'tab_CR'}
\begin{tabular}{|l|l|c|}
\hline
Containing BB (ET) & Contained BB (Abn.) & CR\\
\hline
GT & Radiologists' Annotations & 74.84\\
GT & YOLO-v5 \cite{yolov5} Pred & 25.67 \\
Pred & YOLO-v5 \cite{yolov5} Pred & 27 \\
GT & MedRPG \cite{medRPG} Pred & 47.55 \\
Pred & MedRPG \cite{medRPG} Pred & 56.14 \\
\hline
\end{tabular}
\end{table}

\section{Discussion and Conclusion}\label{discussion}
We proposed an automatic pipeline to generate BBs for report sentences, from timestamped ET data and report. Despite limited datasets, clinicians can efficiently produce this data during image interpretation. The extracted BBs can guide a DNN to focus on informative image regions and assess the model explainability by calculating containment ratio.

We illustrated that the phrase grounding model outperformed the object detection model; however, phrase grounding can benefit from some specifications of object detection models. For instance, predicting multiple bounding boxes for each statement perfectly matches the the nature of medical domain and the confidence score corresponds to the certainty level of the radiologists regarding abnormalities.

One of the main limitations of this study is the small size of the dataset; applying these approaches to a bigger dataset will produce more reliable results. The other limitation was the lack of a radiologist collaborator to approve statements we extracted for bounding boxes in dataset repurposing.

Incorporating multiple bounding box prediction associated with confidence scores and integrating ET into a phrase grounding model to enhance its performance and explainability are the future directions of this study.

\section{Complience with Ethical Standards}
MIMIC-CXR and REFLACX are publicly available datasets.

\bibliographystyle{IEEEbib}
\bibliography{strings}

\end{document}